\documentclass[10pt,twocolumn,letterpaper]{article}

\usepackage[pagenumbers]{cvpr} % To force page numbers, e.g. for an arXiv version

\definecolor{cvprblue}{rgb}{0.21,0.49,0.74}
\usepackage[pagebackref,breaklinks,colorlinks,allcolors=cvprblue]{hyperref}
\usepackage{bm}
\usepackage{stfloats}
\usepackage{colortbl}
\usepackage{booktabs}
\usepackage{multirow}
\usepackage{booktabs}
\usepackage{pifont}
\usepackage{dsfont}
\usepackage{cuted}

\newcommand{\cmark}{\ding{51}}  % ✓
\newcommand{\xmark}{\ding{55}}  % ✗

\usepackage[most]{tcolorbox}
\usepackage{float}
\usepackage{xspace}
\tcbset{
  aibox/.style={
    width=474.18663pt,
    top=10pt,
    colback=white,
    colframe=black,
    colbacktitle=black,
    enhanced,
    center,
    attach boxed title to top left={yshift=-0.1in,xshift=0.15in},
    boxed title style={boxrule=0pt,colframe=white,},
  }
}
\newtcolorbox{AIbox}[2][]{aibox,title=#2,#1}

\def\paperID{38380} % *** Enter the Paper ID here
\def\confName{CVPR}
\def\confYear{2026}

\title{Omni-MMSI: Toward Identity-attributed Social Interaction Understanding}

\author{
Xinpeng Li$^{1}$ \quad
Bolin Lai$^{2}$ \quad
Hardy Chen$^{3}$ \quad
Shijian Deng$^{1}$ \\
Cihang Xie$^{3}$ \quad
Yuyin Zhou$^{3}$ \quad
James M. Rehg$^{4}$ \quad
Yapeng Tian$^{1}$\\
$^{1}$University of Texas at Dallas \quad
$^{2}$Georgia Institute of Technology \\
$^{3}$University of California, Santa Cruz \quad
$^{4}$University of Illinois Urbana-Champaign \\
{\tt\small\{xinpeng.li, shijian.deng, yapeng.tian\}@utdallas.edu} \\
\tt\small{bolin.lai@gatech.edu} \quad 
\tt\small{\{hchen403, cixie, yzhou284\}@ucsc.edu} \quad
\tt\small{jrehg@illinois.edu}
}

\begin{document}
\maketitle
\begin{abstract}
We introduce \textbf{Omni-MMSI}, a new task that requires comprehensive social interaction understanding from raw audio, vision, and speech input. 
The task involves perceiving identity-attributed social cues (e.g., who is speaking what) and reasoning about the social interaction (e.g., whom the speaker refers to).
This task is essential for developing AI assistants that can perceive and respond to human interactions.
Unlike prior studies that operate on oracle-preprocessed social cues, Omni-MMSI reflects realistic scenarios where AI assistants must perceive and reason from raw data.
However, existing pipelines and multi-modal LLMs perform poorly on Omni-MMSI because they lack reliable identity attribution capabilities, which leads to inaccurate social interaction understanding.
To address this challenge, we propose \textbf{Omni-MMSI-R}, a reference-guided pipeline that produces identity-attributed social cues with tools and conducts chain-of-thought social reasoning. 
To facilitate this pipeline, we construct participant-level reference pairs and curate reasoning annotations on top of the existing datasets.
Experiments demonstrate that Omni-MMSI-R outperforms advanced LLMs and counterparts on Omni-MMSI.
Project page: \url{https://sampson-lee.github.io/omni-mmsi-project-page}.
\end{abstract}

\section{Introduction}
\label{sec:intro}

\begin{figure}[t]
    \centering
    \includegraphics[width=1.0\linewidth]{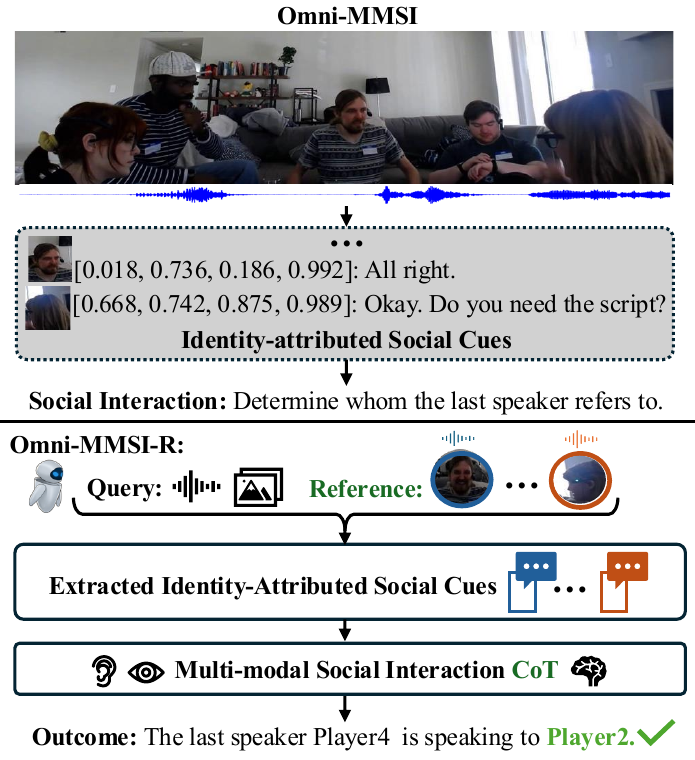}
    \caption{Overview of the Omni-MMSI task and Omni-MMSI-R pipeline. The Omni-MMSI explores social interaction understanding in a multi-party social scene only using raw audio and video, unlike prior studies that assume identity-attributed social cues are perfectly provided. To address the challenge of attribution, our Omni-MMSI-R is explicitly guided by individual references to generate identity-attributed multi-modal cues and performs CoT reasoning for accurate social interaction understanding.}

    \label{fig:omni-intrudction}
    \vspace{-5mm}
\end{figure}

Multi-modal Multi-party Social Interaction Understanding (MMSI), aiming to interpret human behaviors in social situations, is fundamental for advancing socially-intelligent AI systems~\citep{lee2024towards, lai2023werewolf, lee2024modeling, li2024socialgpt, Feng2025TMLR-SocialAgents}. As shown in \Cref{fig:omni-intrudction}, given audio-video input, the system is required to extract identity-attributed verbal and non-verbal social cues. For instance, the chronological utterances, \textit{[Player2]: All right. [Player4]: Okay. Do you need the script?}, and their corresponding bounding boxes, \textit{[0.018, 0.736, 0.186, 0.992] and [0.668, 0.742, 0.875, 0.989]}, constitute essential identity-attributed social cues. Then, the system should analyze these multi-modal social cues to infer the social interaction, \ie \textit{determine whom the last speaker refers to in the query audio-video}. These capabilities are essential for enabling AI assistants that can perceive, reason over, and respond to human interactions in social scenarios~\citep{haber2020making, elsherbini2023towards, breazeal2016social}.

Recent computer vision studies have explored social interaction understanding and advanced it with representation alignment~\cite{lee2024modeling} and conversation forecasting~\cite{li2025towards}. Despite the rapid progress, they remain limited in scope: they assume the individual-attributed social cues are perfectly provided, typically via oracle-preprocessing. However, in real-world deployment, AI assistants must understand social interactions from raw data input. To better align with realistic applications, we introduce a new task, named \textbf{Omni-MMSI}, which requires social interaction understanding on raw audio-video input. The system needs to extract identity-attributed social cues, including who speaks what and where they are, and then infer the social interaction.

However, identity attribution is challenging in multi-party scenes, where people show subtle movements, and their voices also sound alike, with a lot of overlap. First, the off-the-shelf extractors~\cite{radford2023robust, jocher2022ultralytics} that can be used in earlier studies were designed for single-person scenarios and fail to handle the crucial attribution step required in Omni-MMSI. Second, while Omni-modal Large Language Models (Omni-LLMs) demonstrate strong cue extraction, they still struggle to correctly associate these cues with individuals across modalities. Therefore, prior pipelines and Omni-LLMs degrade significantly when transitioning from oracle identity-attributed cues to raw inputs. As shown in \cref{fig:omni-challenge}, the accuracy of prior pipelines~\cite{lee2024modeling, li2025towards} drops by an average of 28.1\%, and even human annotators and advanced Omni-LLMs~\cite{xu2025qwen2, comanici2025gemini} exhibit an average decline of 9.52\%. 

To tackle this challenge, we propose \textbf{Omni-MMSI-R}, a LLM-based pipeline that utilizes references to guide identity attribution. Our key insight is that humans remember the appearance and voice of familiar people, and readily associate their gestures or speech with these memories when interpreting social interactions. In practical use, these references are usually easy to collect on devices through the enrollment or verification processes~\cite{jiang2023target, clarke2025speaker}. As shown in \cref{fig:omni-intrudction}, to generate accurate identity-attributed social cues, task-specific tools associate cues with references. Then, to further enhance MMSI ability, the model performs chain-of-thought (CoT) reasoning. To facilitate such a pipeline, we manually construct paired image-audio references for each sample and curate a CoT reasoning dataset. 

We evaluate Omni-MMSI-R on two social interaction tasks across two social datasets, Ego4D and YouTube~\cite{lee2024modeling}. Our method outperforms previous studies by 12\% on Ego4D and 15.1\% on YouTube in social interaction understanding and exceeds advanced LLMs by 23.7\% on Ego4D and 18.9\% on YouTube in identity attribution, demonstrating that Omni-MMSI-R benefits from reference guidance.

In summary, our contributions are four-fold:
\begin{itemize}
    \item We present Omni-MMSI, a new task for realistic scenarios that requires multi-party multi-modal social interaction understanding only using raw audio-vision input. 
    \item We propose Omni-MMSI-R, a reference-guided pipeline that generates identity-attributed social cues with tools and performs CoT reasoning for accurate MMSI. 
    \item We curate paired audio-vision references and CoT reasoning annotations for two current datasets for future study. 
    \item Experiments on two social interaction tasks across two datasets demonstrate that the proposal benefits from reference guidance and achieves state-of-the-art performance.
\end{itemize}

\section{Related Works}

\subsection{Multi-modal Social Interaction Understanding}

MMSI aims to interpret complex interactions among multiple participants by using verbal and non-verbal cues~\cite{li2024socialgpt, gupta2024mtgs, lee2024towards, lai2023werewolf, cao2025socialgesture, lee2024modeling, li2025towards, zadeh2019social, ouyang2025multi}. 
The non-verbal social cues include visual behaviors such as body gestures, gaze patterns, and facial expressions~\cite{jindal2023contrastive, chong2020detecting, grauman2022ego4d, tafasca2023childplay, benitez2021ipn, liu2021imigue, chen2023smg, kapitanov2024hagrid, zhang2021relative, savchenko2023facial, zhao2024err, li2024two, cao2025toward, mao2025facial, li2021sequential, ryan2025gaze, wei2024nonverbal, nakamura2023deepoint, tonini2023object}. The verbal social cues include  linguistic signals such as conversational dynamics, speaker intent, speaker diarization, and dialogue sentiment~\cite{Feng2025TMLR-SocialAgents, feng2021emowoz, malhotra2022speaker, chen2023benchmark, raman2022conflab, ryan2023egocentric, hou2024multi, cheng2024emotion, lian2025affectgpt, lian2024ov, hyun2023smile, guosns, xu2022ava, liao2023light, park2022review, mingote2024audio}. 

Despite these advances, these works all assume perfectly provided individual-attributed cues as model input, overlooking the gap between raw audio-visual input and attributed social cues in realistic deployment. In contrast, Omni-MMSI focuses on social interaction understanding only using streaming audio and video, where the system must first extract identity-attributed verbal and non-verbal social cues and then reason about the social interaction.

\subsection{Multi-modal Foundation and Reasoning Model}
Multi-modal foundation models pave the way toward better intelligent systems. While proprietary models~\citep{hurst2024gpt,ClaudeAI,team2024gemini} often showcase strong performance, open-weight models~\citep{Qwen2.5-VL,vteam2025glm45vglm41vthinkingversatilemultimodal,kimiteam2025kimivltechnicalreport,grattafiori2024llama3herdmodels, liu2023visual, zhang2023video, zhu2023minigpt, zhang2023internlm, chen2023minigpt, lin2023video, chen2024internvl, cheng2024videollama, zhang2024internlm, thawakar2025llamav, Zhang2024TMLR-MultimodalCoT, xu2025qwen2, abouelenin2025phi, ye2025omnivinci, zhao2025humanomni, zhao2025r1} provide more opportunities for specialized downstream tasks, making them useful for multi-modal social interaction understanding.
%reasoning
Reasoning~\citep{wei2022chain} as an emergent ability of LLMs~\citep{wei2022emergent} has recently attracted attention recently for its effectiveness under text-only settings~\citep{jaech2024openai,guo2025deepseek,tinyzero,openai2025gptoss120bgptoss20bmodel}.
Multi-modal reasoning models extend this success to general image understanding~\citep{xu2024llavacot,deng2025openvlthinkercomplexvisionlanguagereasoning,chen2025sftrlearlyinvestigation,liu2025noisyrolloutreinforcingvisualreasoning,shen2025vlmr1stablegeneralizabler1style,chen2025r1v}, video understanding~\citep{feng2025videor1reinforcingvideoreasoning,zhang2025thinkingvideosmultimodaltoolaugmented,li2025videochatr1enhancingspatiotemporalperception,zhang2025tinyllavavideor1smallerlmmsvideo,wang-etal-2025-video} and some vertical domains like medical image understanding~\citep{huang2025medvlsynthersynthesizinghighqualityvisual,su2025gmai}.
Tooling further extends LLMs' ability to perform a broad spectrum of tasks through the use of tools~\cite{fan2025tool, liu2024llava, gao2024clova, zhang2025deep, zhouvideoagent, gao2024multi, gao2025mmat}.

% OUR WORK
However, CoT reasoning and tooling paradigms remain unexplored in MMSI. To advance computer vision and social AI community, we curate paired audio-vision references and CoT reasoning traces on top of existing datasets, and demonstrate the effectiveness of CoT and tooling.

\begin{figure*}[!t]
    \centering
    \includegraphics[width=\linewidth]{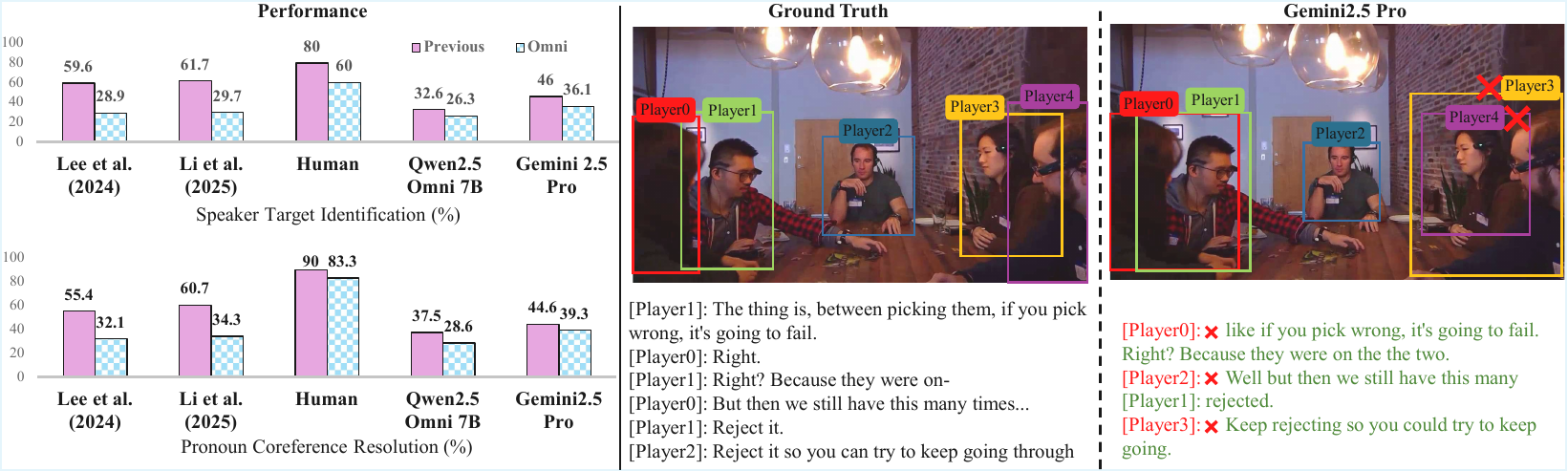}
    \caption{Illustration of the challenge in Omni-MMSI.
    The quantitative results (left) show prior pipelines, humans, and advanced Omni-LLMs show substantial accuracy drops when transitioning from oracle cues to raw audio-video input.
    Typical attribution failures (right), where speech and bounding boxes are mismatched to identities, reveal the weak multi-modal identity attribution of advanced Omni-LLMs.
    }
    \label{fig:omni-challenge}
    % \vspace{-4mm}
\end{figure*}

\section{Problem Formulation and Challenges}

Omni-MMSI pursues the MMSI abilities on raw audio-visual input instead of relying on oracle cues. Specifically, we study two typical MMSI tasks~\cite{lee2024modeling, li2025towards}: Speaking Target Identification (STI) and Pronoun Coreference Resolution (PCR). STI aims to identify who the speaker is talking to when the utterance contains a second-person reference, e.g., ``you'' and ``your''; PCR focuses on resolving which participant a third-person pronoun refers to, e.g., ``he'', ``she'', ``him'', ``her'' and ``his''. The inputs are a raw audio-video segment $I_{AV}$ and system prompt $P$ that configures a specific task. The output $X_{answer}$ is the predicted referent identity. The Omni-MMSI is to build a system $f$:
\begin{equation}
    f : (P, I_{AV}) \rightarrow X_{answer}.
\label{eq:ori_pipeline}
\end{equation}

Unlike previous studies~\cite{li2025towards, lee2024modeling} that assume oracle-preprocessing social cues as input, Omni-MMSI operates on the raw audio-video segment, requiring models to automatically extract social cues and infer social interaction. To assess the challenge, we evaluate performance on the social Ego4D dataset across two social tasks when transferring from oracle input to raw-data one. As shown in~\cref{fig:omni-challenge}, previous pipelines~\citep{lee2024modeling, li2025towards} and advanced Omni-LLMs such as Qwen2.5 Omni 7B~\cite{xu2025qwen2} and Gemini 2.5 Pro~\citep{comanici2025gemini} exhibit significant performance drops, confirming that the Omni-MMSI poses a significant challenge. It also underscores that current LLMs still fall short of human-level understanding in multi-modal and multi-party social reasoning~\citep{inoue2025llm, tan2023chatgpt}.

The major bottleneck is identity attribution ability on raw audio-visual input. On the one hand, off-the-shelf extractors in prior pipelines~\citep{lee2024modeling, li2025towards} are designed for single-person scenarios, failing to attribute cues to individuals in a multi-party setting. On the other hand, although recent Omni-LLMs~\cite{xu2025qwen2, comanici2025gemini} have shown promising performance in extracting cues, they still struggle to associate detected cues with the corresponding subjects. As illustrated in \cref{fig:omni-challenge}, Gemini 2.5 Pro often assigns speech content or bounding boxes to the wrong identity. Specifically, for visual attribution, Gemini 2.5 Pro attributes participants based on their left-to-right spatial order, but this assumption leads to identity swaps when detection fails under occlusion or overlapping. For speech attribution, Gemini 2.5 Pro often mismatches the recognized utterance with wrong identity. Such weak multi-modal association results in inaccurate social cues, ultimately degrading social interaction reasoning.

\section{Methodology}

\begin{figure*}[t]
    \centering
    \includegraphics[width=\linewidth]{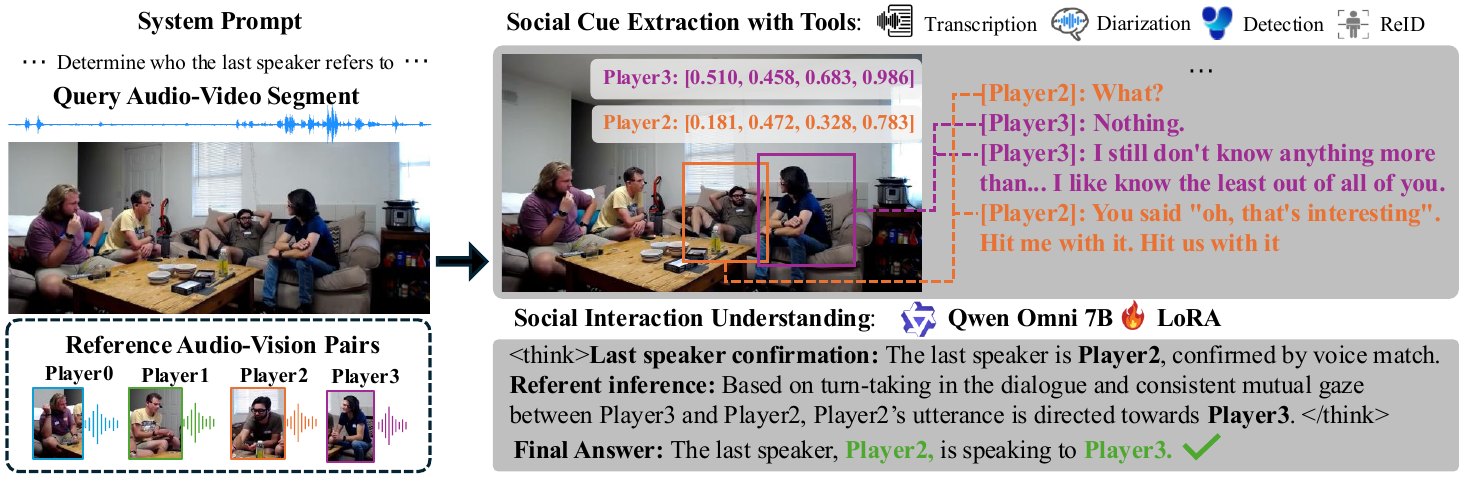}
    \caption{Overview of the Omni-MMSI-R pipeline. Given a query audio-video segment with multiple participants, the system first retrieves reference audio-vision pairs that represent each individual. Task-specific tools, for transcription, diarization, detection and ReID, generate identity-attributed verbal and non-verbal social cues, specifying who speaks what and where they are. These cues, together with the references and the raw audio-video stream, form the reference-guided input. The Omni-LLM (Qwen2.5 Omni 7B fine-tuned with LoRA) then performs chain-of-thought reasoning over this input to produce an accurate response for social interaction understanding.}
    \label{fig:omni-pipeline}
    \vspace{-1mm}
\end{figure*}

\subsection{Overview of Omni-MMSI-R}

To tackle the difficulty of social cues attribution, we introduce Omni-MMSI-R that leverages references $\mathcal{R}$ to generate identity-attributed social cues and perform CoT social reasoning. The system target can be formulated as:
\begin{equation}
    f : (P, I_{AV}, \mathcal{R}) \rightarrow X_{answer}.
\label{eq:ref_pipeline}
\end{equation}

As shown in \cref{fig:omni-pipeline}, given a query audio-video segment, Omni-MMSI-R loads a set of reference audio-image pairs that store representative visual and acoustic profiles for each individual. Based on these references, task-specific tools generate identity-attributed multi-modal social cues, such as conversation transcripts and individual locations. Then, an Omni-LLM performs CoT reasoning on the audio-video segment and reference audio-image pairs, along with generated attributed cues, and produces an accurate answer.

\subsection{Reference Guidance}
To address the difficulty of identity attribution on raw audio-video input, we propose to associate social cues with guided references. The insight is that humans rely on the appearance and voice of memorized people to guide identity association in multi-party situations. In practical use, the references are usually easy to collect on devices through the enrollment or verification processes~\cite{jiang2023target, clarke2025speaker}.

\begin{figure}[t]
    \centering
    \includegraphics[width=\linewidth]{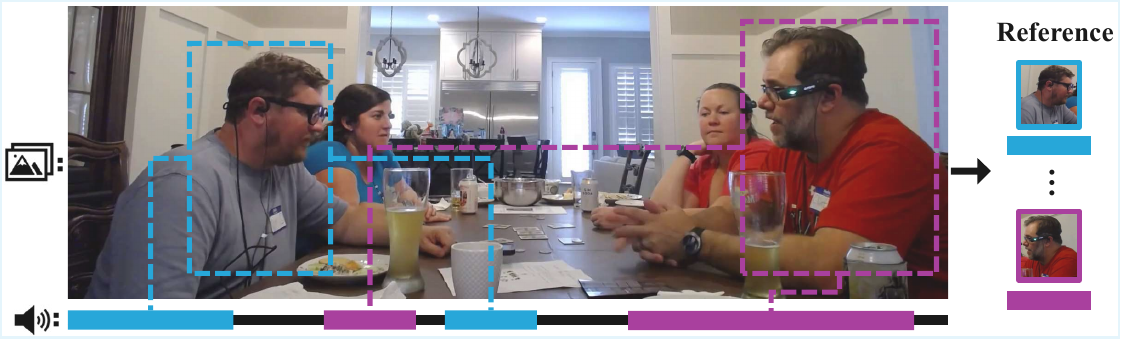}
    \caption{Illustration of preparation of reference audio-vision pairs for each participant, which serve as anchors for identity attribution.}
    \label{fig:omni-reference}
    \vspace{-6mm}
\end{figure}

For research purposes, we manually crop each participant’s upper body image and extract several corresponding voice clips to build the reference pairs, as shown in \Cref{fig:omni-reference}. In total, we curate 69 audio-visual reference profiles covering different participants across the experimental datasets.

Omni-MMSI-R can access the reference audio-visual set
\(\mathcal{R} = \{(a_i, v_i)\}_{i=1}^{N}\)
for all \(N\) participants in the scene, where \(a_i\) and \(v_i\) denote the representative voice and appearance of participant \(i\). These references anchor identities across modalities and time, reducing common failures, like identity swaps under occlusion and cross-modal mismatches, and yielding accurate identity-attributed social cues.

\subsection{Social Cue Extraction with Tools}
To generate accurate cues, we leverage tools to help detect social cues and associate them with reference identities.

\noindent\textbf{Audio Tools.} 
We first apply Whisper~\cite{radford2023robust} to transcribe the query audio into a sequence of utterances with timestamps. 
For each utterance, SpeechBrain~\cite{ravanelli2021speechbrain} performs speaker verification by encoding both the utterance audio and each reference voice into embeddings and computing their cosine similarity. 
The reference with the highest similarity is selected as the predicted speaker identity. 
This process yields identity-attributed verbal social cues that contain transcribed speech and the corresponding speaker identity.

\noindent\textbf{Visual Tools.} 
We first leverage YOLO~\cite{jocher2022ultralytics} to detect all visible participants in the last frame of the query video. 
For every detected bounding box, we then employ OSNet~\cite{zhou2019omni} for person re-identification. 
Specifically, both the detected image crop and each reference image are encoded into visual embeddings, and the similarity between them is computed. 
The reference with the highest similarity is selected as the predicted visual identity. 
This produces identity-attributed non-verbal social cues that specify both the spatial position and identity of each participant in the scene.

After extraction, the identity-attributed social cues \(\mathcal{S}\), along with the query audio-video segment \(I_{AV}\) and the reference audio-image pairs \(\mathcal{R}\), are fed into an Omni-LLM. 

\subsection{Social Interaction Understanding with CoT}
Omni-MMSI naturally involves multi-step fine-grained understanding: (1) confirming the last speaker from audio, visual, and speech evidence, and (2) inferring the speaker’s referent by integrating verbal cues, such as matching the utterance with prior dialog and speaker context, and non-verbal interaction signals, including mutual eye contact or pointing. Training models to directly output answers often fails to capture the fine-grained evidence, resulting in less reliable responses. Therefore, we propose to supervise the model with structured CoT reasoning traces.

\begin{figure}[t]
    \centering
    \includegraphics[width=\linewidth]{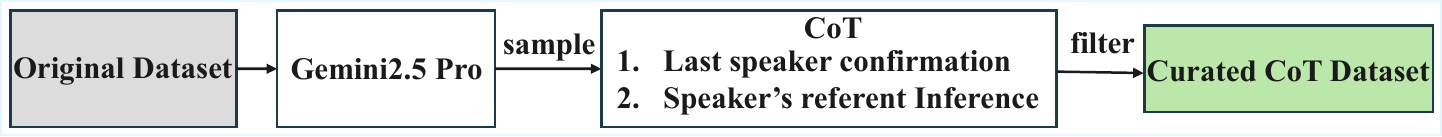}
    \caption{Illustration of the construction of CoT datasets.}
    \label{fig:omni-cot}
\vspace{-6mm}
\end{figure}

\noindent\textbf{CoT Data Curation.} To facilitate the training, we curate CoT annotations by a generate-and-filter pipeline. As illustrated in \cref{fig:omni-cot}, (i) we upload query segment, reference input, and social cues to Gemini 2.5 Pro and request it to generate both social reasoning traces and a final answer, including last speaker confirmation and referent inference with verbal and non-verbal evidence. (ii) Based on the rejection sampling principle, a generated sample is retained only if the reasoning trace leads to a final answer that is consistent with the ground truth. Otherwise, we repeatedly query Gemini~2.5~Pro until a correct answer is obtained, or stop after 10 attempts. (iii) To further ensure the CoT quality, we perform a lightweight human review to discard or minimally revise reasoning traces that are implausible or inconsistent with the audio-visual evidence. Through this process, we obtain a set of samples with reliable and interpretable social reasoning traces. \cref{fig:omni-pipeline} shows a CoT example: \textit{\textless think\textgreater Last speaker confirmation: The last speaker is Player2, confirmed by voice match. Speaker's referent inference: Based on turn-taking in the dialogue and consistent mutual gaze between Player3 and Player2, Player2’s utterance is directed towards Player3. \textless/think\textgreater}

\noindent\textbf{Model Training.} After obtaining the CoT reasoning traces $X_{think}$, we train the model for MMSI, formulated as:

\begin{equation}
    X_{answer}, X_{think} \;=\; f_{\theta}^{\text{Omni-LLM}}\bigl(P, I_{AV},\mathcal{R},\mathcal{S}\bigr),
\label{eq:final_pipeline}
\end{equation}

\noindent where \(f_{\theta}^{\text{Omni-LLM}}\) denotes the Omni-LLM. By learning reasoning over raw data, augmented with references and tool-extracted cues, the model can address Omni-MMSI.

\section{Experiments}

\subsection{Implementation Details}

We select \textbf{Qwen2.5-Omni-7B}~\citep{xu2025qwen2} as our omni-modal large language model and {LLaMA-Factory}~\cite{zheng2024llamafactory} framework for supervised fine-tuning (SFT).
We apply LoRA~\citep{hu2022lora} fine-tuning with a rank of 8 while other LoRA hyperparameters follow LLaMA-Factory defaults.
Training uses cross-entropy loss, a cosine learning-rate scheduler with 10\% warm-up and a context length of 16,384 tokens. We train for 3 epochs with per-device batch size 1 and gradient accumulation 1. The learning rate is set to \(1 \times 10^{-4}\) empirically for the speaking target identification and the pronoun coreference resolution task.
The query segment is standardized to contain 5 dialogue turns, with average duration of 14 seconds.
The reference audio clips are trimmed to 5 seconds, whereas the reference images vary in size.
Additional implementation details are provided in the supplementary.

\begin{figure*}[t]
    \centering
    \includegraphics[width=\linewidth]{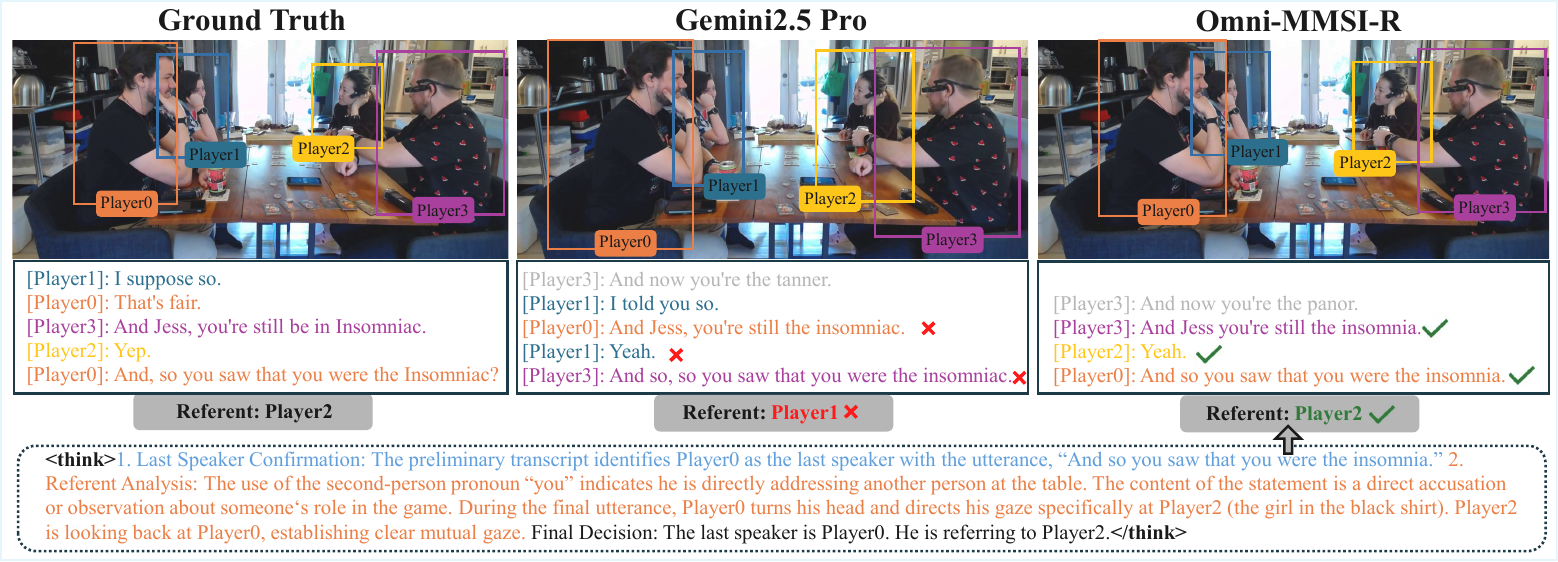}
    \caption{Qualitative comparison between Gemini2.5 Pro and our proposed Omni-MMSI-R on multi-party situations.  Gemini2.5 Pro often misattributes utterances to incorrect visual identities, leading to wrong referent predictions, while Omni-MMSI-R accurately aligns verbal and non-verbal cues with individual references, yielding reliable identity-attributed social cues for social interaction reasoning.}
    \label{fig:exp_qualitative}
    \vspace{-3mm}
\end{figure*}

\subsection{Dataset and Metrics}
Experiments are conducted on the Werewolf Among Us dataset, which comprises two subsets (YouTube and Ego4D) of social deduction games~\citep{lai2023werewolf}. We follow~\citet{lee2024modeling} and ~\citet{li2025towards} to set up STI and PCR tasks. In Omni-MMSI, we remove oracle cues, transcript and keypoints, but provide references and CoT reasoning traces.

\noindent\textbf{YouTube} contains 3,255 samples for STI and 2,679 samples for PCR, with an average of 5 individuals per sample. For each sample, we manually construct reference audio-image pairs. For the training split, we generate CoT reasoning traces and filter out 2,124 samples for STI (average 202 words) and 1,935 samples for PCR (average 220 words).

\noindent\textbf{Ego4D} contains 832 STI samples and 503 PCR samples, with each sample involving an average of 5 individuals. For each sample, we manually construct reference audio-image pairs. For the training split, we generate CoT reasoning traces and filter out 521 samples for STI (average 206 words) and 321 samples for PCR (average 226 words).

\noindent\textbf{Evaluation.}
To evaluate social interaction understanding, we follow previous studies~\cite{lee2024modeling, li2025towards} to report the overall accuracy of the predicted referent. To further evaluate identity attribution ability, we compute the accuracy of attributed identity of each utterance (verbal attribution) and detected location on the last frame (non-verbal attribution).

\begin{table}[t]
\centering
\caption{Performance comparison of different pipelines on STI and PCR tasks. The upper block reports results (\%) on Ego4D, and the lower on YouTube. The results highlight the effectiveness of our Omni-MMSI-R pipeline design for Omni-MMSI.}
\begin{tabular}{lccc}
\toprule
\textbf{Pipeline (Ego4D)} & \textbf{STI} & \textbf{PCR} & \textbf{Avg. Acc.} \\
\midrule
Qwen2.5 Omni 7B~\citep{xu2025qwen2} & 26.29 & 28.57 & 27.43 \\
Phi-4-Multimodal~\cite{abouelenin2025phi} & 14.86 & 8.93 & 11.90 \\
HumanOmni~\cite{zhao2025humanomni} & 21.71 & 14.29 & 18.00 \\
R1-Omni~\cite{zhao2025r1} & 0.57 & 0.00 & 0.29 \\
OmniVinci~\cite{ye2025omnivinci} & 24.57 & 37.50 & 31.04 \\
Qwen3 Omni 30B~\cite{xu2025qwen3} & 30.86 & 28.57 & 29.72 \\
Gemini 2.5 Pro~\cite{comanici2025gemini} & 36.12 & 39.28 & \underline{37.70} \\
Lee et al.*~\cite{lee2024modeling} & 28.98 & 32.14 & 30.56 \\
Li et al.*~\cite{li2025towards} & 29.73 & 32.27 & 31.00 \\
\rowcolor{gray!15}
\textbf{Omni-MMSI-R (ours)} & \textbf{40.57} & \textbf{45.54} & \textbf{43.06} \\
\midrule
\textbf{Pipeline (YouTube)} & \textbf{STI} & \textbf{PCR} & \textbf{Avg. Acc.} \\
\midrule
Qwen2.5 Omni 7B~\citep{xu2025qwen2} & 14.00 & 26.18 & 20.09 \\
Phi-4-Multimodal~\cite{abouelenin2025phi} & 16.21 & 21.11 & 18.66 \\
HumanOmni~\cite{zhao2025humanomni} & 20.80 & 28.02 & 24.41 \\
R1-Omni~\cite{zhao2025r1} & 0.15 & 0.19 & 0.17 \\
OmniVinci~\cite{ye2025omnivinci} & 19.57 & 29.17 & 24.37 \\
Qwen3 Omni 30B~\cite{xu2025qwen3} & 18.81 & 36.85 & 27.83 \\
Gemini 2.5 Pro~\cite{comanici2025gemini} & 36.13 & 53.47 & \underline{44.80} \\
Lee et al.*~\cite{lee2024modeling} & 29.01 & 34.80 & 31.91 \\
Li et al.*~\cite{li2025towards} & 26.30 & 30.14 & 28.22 \\
\rowcolor{gray!15}
\textbf{Omni-MMSI-R (ours)} & \textbf{37.46} & \textbf{56.62} & \textbf{47.04} \\
\bottomrule
\end{tabular}
\label{tab:exp_siu}
% \vspace{-8mm}
\end{table}
\subsection{Pipeline Performance Comparison}

To evaluate the effectiveness of different pipelines for~\eqref{eq:ori_pipeline}, we conduct evaluation on Ego4D and YouTube. For recent advanced Omni-LLMs~\cite{xu2025qwen2, abouelenin2025phi, zhao2025humanomni, zhao2025r1, ye2025omnivinci, xu2025qwen3, comanici2025gemini}, we directly feed them with the query audio-video pairs and prompt them to generate attributed social cues and social interaction answers. Note that participant identities are deterministically defined by spatial ordering in the system prompt. For the previous MMSI counterparts~\cite{lee2024modeling, li2025towards}, which overlook the attribution process, we first generate unattributed social cues using extractors and then feed them to the model. 

\cref{tab:exp_siu} quantitatively compares the pipelines on social interaction understanding, including STI and PCR. Omni-MMSI-R achieves state-of-the-art performance, reaching 43.06\% on Ego4D and 47.04\% on YouTube.
Relative to existing Omni-LLMs, Omni-MMSI-R improves the average accuracy by 5.36\% on Ego4D and 2.24\% on YouTube. Compared to previous MMSI pipelines, the improvement reaches 12.06\% on Ego4D and 15.13\% on YouTube.
These results confirm that explicit reference guidance greatly strengthens multi-modal social interaction reasoning.

Beyond interaction reasoning, \cref{tab:exp_attr} reports evaluation results on social cues attribution, including verbal and non-verbal attribution accuracy. Omni-MMSI-R substantially outperforms strong Omni-LLMs, improving the average attribution accuracy by 23.68\% on Ego4D and 18.91\% on YouTube. Note that since some weak Omni-LLMs~\cite{xu2025qwen2, abouelenin2025phi, zhao2025humanomni, zhao2025r1} fail to generate valid identity-attributed social cues during inference, their attribution accuracy is not reported. These improvements indicate that references significantly enhance the pipeline's ability of identity attribution.

\cref{fig:exp_qualitative} illustrates qualitative comparisons between Gemini 2.5 Pro and our proposed Omni-MMSI-R. We can see Gemini 2.5 Pro fails to attribute utterances to the right visual identities, leading to inaccurate referent prediction. It reflects its limited ability in cross-modal attribution for multi-modal social interaction understanding. Instead, Omni-MMSI-R correctly aligns verbal and non-verbal cues to individual references, producing more reliable identity-attributed social cues. Based on these cues, the model performs CoT reasoning with last speaker confirmation and referent analysis to obtain an accurate social interaction answer. Overall, these quantitative and qualitative results validate the effectiveness of our reference-guided pipeline.

\begin{table}[t]
\centering
\caption{Performance comparison of different pipelines on social cues attribution, including Verbal, Non-Verbal, and Average Attribution accuracy (\%). The upper block reports results on Ego4D, and the lower block reports results on YouTube. The results show our Omni-MMSI-R pipeline can achieve better identity attribution.}
\begin{tabular}{lccc}
\toprule
\textbf{Pipeline (Ego4D)} & \textbf{Verbal} & \textbf{Non-Verbal} & \textbf{Avg.} \\
\midrule
OmniVinci~\cite{ye2025omnivinci} & 54.04 & 27.42 & 40.73 \\
Qwen3 Omni 30B~\cite{xu2025qwen3} & 52.61 & 57.61 & \underline{55.11} \\
Gemini 2.5 Pro~\cite{comanici2025gemini} & 44.75 & 26.52 & 35.64 \\
\rowcolor{gray!15}
\textbf{Omni-MMSI-R (ours)} & \textbf{71.09} & \textbf{86.48} & \textbf{78.79} \\
\midrule
\textbf{Pipeline (YouTube)} & \textbf{Verbal} & \textbf{Non-Verbal} & \textbf{Avg.} \\
\midrule
OmniVinci~\cite{ye2025omnivinci} & 58.10 & 32.35 & 45.23 \\
Qwen3 Omni 30B~\cite{xu2025qwen3} & 57.73 & 52.54 & 55.14 \\
Gemini 2.5 Pro~\cite{comanici2025gemini} & 50.57 & 65.51 & \underline{58.04} \\
\rowcolor{gray!15}
\textbf{Omni-MMSI-R (ours)} & \textbf{67.57} & \textbf{86.33} & \textbf{76.95} \\
\bottomrule
\end{tabular}
\label{tab:exp_attr}
\vspace{-6mm}
\end{table}

\subsection{Referential Pipeline Comparison}
To evaluate the effectiveness of different pipelines for~\eqref{eq:ref_pipeline}, we compare our proposal with Omni-LLMs~\cite{xu2025qwen2, ye2025omnivinci, xu2025qwen3, comanici2025gemini} on Ego4D and YouTube. We provide the references along with query videos to Omni-LLMs for social cue attribution and social interaction understanding. For social interaction understanding, as shown in~\cref{tab:exp_siu_ref}, Omni-MMSI-R achieves comparable accuracy to Gemini~2.5~Pro and exceeds open-source Omni-LLMs by 9.12\% on Ego4D and 11.56\% on YouTube. Compared to non-reference setting, large Omni-LLMs like Qwen3~Omni~30B~\cite{xu2025qwen3} and Gemini~2.5~Pro~\cite{comanici2025gemini} obtain performance gains, demonstrating the benefits of reference guidance. However, small Omni-LLMs like Qwen2.5~Omni~7B~\citep{xu2025qwen2} and OmniVinci~\cite{ye2025omnivinci} degrade in performance. Small models might not be able to utilize the reference, showing the necessity of using tools.

For identity attribution, \cref{tab:supp_reference} shows reference guidance generally improves attribution for Gemini~2.5~Pro, which achieves gains of 11.53\% on Ego4D and 4.99\% on YouTube. However, not all Omni-LLMs can reliably incorporate references. OmniVinci~\cite{ye2025omnivinci} cannot produce valid social cues when receiving both query and reference audio-vision pairs, so its attribution accuracy cannot be reported. Qwen3 Omni~30B~\cite{xu2025qwen3} shows lower attribution accuracy after including reference pairs. This indicates LLMs alone struggle to use references effectively for identity attribution. In addition, generating identity-attributed verbal and non-verbal cues through LLMs introduces considerable inference latency. In comparison, the lightweight tools in Omni-MMSI-R provide fast and reliable social cues.

\begin{table}[t]
\centering
\caption{Performance comparison of different referential pipelines on STI and PCR tasks. The upper block reports results (\%) on Ego4D, and the lower on YouTube. The results highlight the effectiveness of our Omni-MMSI-R pipeline design for Omni-MMSI.}
\begin{tabular}{lccc}
\toprule
\textbf{Pipeline (Ego4D)} & \textbf{STI} & \textbf{PCR} & \textbf{Avg. Acc.} \\
\midrule
Qwen2.5 Omni 7B~\cite{xu2025qwen2} & 21.23 & 10.71 & 15.97 \\
% Qwen2.5 Omni 7B (SFT) & 32.78 & 39.18 & 35.98 \\
OmniVinci~\cite{ye2025omnivinci} & 24.00 & 36.61 & 30.31 \\
Qwen3 Omni 30B~\cite{xu2025qwen3} & 28.57 & 39.28 & 33.94 \\
Gemini 2.5 Pro~\cite{comanici2025gemini} & 40.57 & 44.64 & \underline{42.61} \\
\rowcolor{gray!15}
\textbf{Omni-MMSI-R (ours)} & \textbf{40.57} & \textbf{45.54} & \textbf{43.06} \\
\midrule
\textbf{Pipeline (YouTube)} & \textbf{STI} & \textbf{PCR} & \textbf{Avg. Acc.} \\
\midrule
Qwen2.5 Omni 7B~\citep{xu2025qwen2} & 12.08 & 15.16 & 13.62 \\
% Qwen2.5 Omni 7B (SFT) & 28.59 & 44.33 & 36.46 \\
OmniVinci~\cite{ye2025omnivinci} & 20.95 & 28.29 & 24.62 \\
Qwen3 Omni 30B~\cite{xu2025qwen3} & 34.78 & 36.19 & 35.48 \\
Gemini 2.5 Pro~\cite{comanici2025gemini} & 39.87 & 57.58 & \textbf{48.72} \\
\rowcolor{gray!15}
\textbf{Omni-MMSI-R (ours)} & 37.46 & 56.62 & \underline{47.04} \\
\bottomrule
\end{tabular}
\label{tab:exp_siu_ref}
% \vspace{-6mm}
\end{table}

\begin{table}[t]
\centering
\caption{Comparison of referential pipelines on social cues attribution, including Verbal, Non-Verbal, and Average Attribution accuracy (\%). The upper block reports results on Ego4D, and the lower block on YouTube. The results show Omni-MMSI-R, using tools, provides the strongest attribution performance.}
\begin{tabular}{lccc}
\toprule
\textbf{Pipeline (Ego4D)} & \textbf{Verbal} & \textbf{N-Verbal} & \textbf{Avg.} \\
\midrule
Qwen3 Omni 30B~\cite{xu2025qwen3} & 21.92 & 71.26 & 46.59 \\
Gemini 2.5 Pro~\cite{comanici2025gemini} & 59.09 & 35.24 & \underline{47.17} \\
\rowcolor{gray!15}
\textbf{Omni-MMSI-R (ours)} & \textbf{71.09} & \textbf{86.48} & \textbf{78.79} \\
\midrule
\textbf{Pipeline (YouTube)} & \textbf{Verbal} & \textbf{N-Verbal} & \textbf{Avg.} \\
\midrule
Qwen3 Omni 30B~\cite{xu2025qwen3} & 23.37 & 50.39 & 36.88 \\
Gemini 2.5 Pro~\cite{comanici2025gemini} & 60.61 & 65.45 & \underline{63.03} \\
\rowcolor{gray!15}
\textbf{Omni-MMSI-R (ours)} & \textbf{67.57} & \textbf{86.33} & \textbf{76.95} \\
\bottomrule
\end{tabular}
\label{tab:supp_reference}
\end{table}

\subsection{Effects of Different Reference-guided Input} 

\noindent
To further investigate how different reference-guided input in~\eqref{eq:final_pipeline} contribute, we conduct ablation studies on Ego4D for social interaction understanding. The baseline means finetuning with only the query audio-video segment. The complete reference-guided input further includes (i) the reference voice-image pairs anchoring individual identities, and (ii) the tool-extracted social cues, consisting of attributed verbal and non-verbal cues. Note that modality is paired: audio references enable verbal cues, while visual references enable non-verbal cues. When one modality is removed, the corresponding attributed cues are also excluded.

As shown in \Cref{tab:ref-input-effect}, the baseline model that finetuned with the query audio-video segment achieves an average accuracy of 33.97\%. Adding audio-vision references without attributed cues improves the performance to 35.98\%, indicating that raw references alone already help ground more reliable social cues implicitly. Adding attributed cues without audio-image pairs boosts the performance to 39.44\%, showing that explicit extracted cues help model understanding social interaction in raw data. By jointly using raw reference data and extracted cues, the model obtains the highest performance 43.05\%. It demonstrates that our LLM is not restricted to blindly trusting extracted cues: (1) The LLM is prompted to jointly use extracted identity cues and direct audio-visual evidence from the video, allowing inaccurate cues to be complemented by raw evidence or corrected. (2) The LLM performs explicit reflection on the last speaker identity in its CoT reasoning. As illustrated in~\cref{fig:supp_example}, the first CoT step verifies the last speaker by jointly examining voice similarity and visible mouth movement. 

Compared to the baseline, incorporating the audio modality together with its attributed verbal cues further increases the accuracy to 39.84\%, while adding the vision modality and its attributed non-verbal cues yields 38.56\%. 
When all modalities and their attributed cues are jointly used, the model achieves the highest accuracy of 43.06\%, demonstrating complementary contribution of different modalities. These results show multi-modal reference and extracted identity-attributed cue together provide the strongest social cues for social interaction reasoning.

\begin{table}[t]
\centering
\caption{Effect of different reference-guided input configurations on social interaction understanding (\%). 
\textit{RA}: Reference Audio, 
\textit{RV}: Reference Visual Image, 
\textit{VC}: Verbal Cues, 
\textit{NC}: Non-Verbal Cues. 
% \cmark{} and \xmark{} indicate the presence or absence.
The results show leveraging audio-visual reference and tool-extracted cue together brings the highest performance.}
\begin{tabular}{ccccccc}
\toprule
RA & RV & VC & NC & STI & PCR & Avg. Acc. \\
\midrule
\xmark & \xmark & \xmark & \xmark & 29.19 & 38.68 & 33.97 \\
\cmark & \cmark & \xmark & \xmark & 32.78 & 39.18 & 35.98 \\
\xmark & \xmark & \cmark & \cmark & 37.14 & 41.75 & 39.44 \\
\cmark & \xmark & \cmark & \xmark & 37.89 & 41.78 & 39.84 \\
\xmark & \cmark & \xmark & \cmark & 34.78 & 42.33 & 38.56 \\
\rowcolor{gray!10}
\cmark & \cmark & \cmark & \cmark & \textbf{40.57} & \textbf{45.54} & \textbf{43.06} \\
\bottomrule
\end{tabular}
\label{tab:ref-input-effect}
\vspace{-2mm}
\end{table}

\subsection{Effectiveness of CoT Reasoning}

To analyze the effectiveness of CoT reasoning, we conduct ablation studies on Ego4D dataset. First, we remove the reference pairs and extracted social cues from the input to examine whether reasoning helps social understanding from raw query input. Since last-speaker confirmation depends on the references, we remove that part and keep only referent inference in the reasoning traces to supervise the model. \cref{tab:reasoning-effect} shows adding only the CoT reasoning enhances performance over the baseline by 1.5\%, indicating that reasoning benefits complex social interaction understanding. This may be because the model is trained with fine-grained evidence grounding contained in the reasoning traces. Therefore, the model can better exploit the multi-modal cues, such as pointing and spoken utterances. When jointly using reference-guided input and CoT reasoning, our model achieves the best performance with an average accuracy of 43.06\%, a significant improvement of 9.1\% compared to baseline, demonstrating their complementary roles: reference-guided input provides more reliable cues, while CoT supervision brings more accurate social interaction understanding. For instance, as you can see in \cref{fig:exp_qualitative}, with accurate reference guidance, the model performs more accurate reasoning to confirm the last speaker as Player0; with CoT, the model exploits fine-grained multi-modal cues.

Second, we investigate the effect of reasoning granularity in CoT supervision, which determines how many intermediate reasoning steps are included during model training. We define four levels of reasoning granularity. The \textit{None} setting provides no intermediate reasoning. The \textit{1-step} setting performs referent inference, where the model explicitly reasons about the target referent of the last speaker. The \textit{2-step} setting further adds last speaker confirmation before referent inference, and the \textit{3-step} setting additionally introduces social cues extraction on top of \textit{2-step}, where the model itself is required to recognize more fine-grained verbal and non-verbal social cues prior to reasoning.  

As shown in \cref{tab:reasoning-effect}, introducing moderate reasoning granularity substantially improves performance. Adding referent inference slightly increases the average accuracy compared with no reasoning, and further including last speaker confirmation yields the highest average accuracy. However, adding one more reasoning stage, explicit social cues extraction, leads to a noticeable decrease in performance. This may be attributed to three factors: first, overly long reasoning sequences could distract the model from focusing on the key reasoning path; second, the model’s limited ability to accurately perceive and utilize social cues makes such explicit cue extraction overly demanding; and third, training data size may be insufficient to support effective learning of such multi-step reasoning processes. Overall, these results demonstrate that a \textit{2-step} CoT supervision strategy, which includes first confirming the speaker and then inferring the referent, achieves the best performance.

\begin{table}[t]
\centering
\caption{Effect of CoT reasoning and different granularity on Ego4D. 
\textit{Reference} indicates whether reference pairs and attributed social cues are provided. 
\textit{Reasoning} controls the level of CoT supervision: 
\textit{None} denotes no reasoning; \textit{CoT} denotes generic reasoning without structured decomposition; 
\textit{1-step}, \textit{2-step}, and \textit{3-step} represent increasingly fine-grained reasoning strategies. 
The results show that CoT improves performance even without reference, while combining reference with structured reasoning yields the best results, with \textit{2-step} achieving the optimal balance.}
\begin{tabular}{c c c c c}
\toprule
Reference & Reasoning & STI & PCR & Avg. Acc. \\
\midrule
\xmark & \xmark        & 29.19 & 38.75 & 33.97 \\
\xmark & \cmark        & 30.71 & 40.18 & 35.45 \\
\midrule
\cmark & None        & 36.57 & 42.25 & 39.41 \\
\cmark & 1-step      & 36.65 & 42.75 & 39.70 \\
\rowcolor{gray!15}
\cmark & 2-step      & \textbf{40.57} & \textbf{45.54} & \textbf{43.06} \\
\cmark & 3-step      & 29.71 & 39.14 & 34.43 \\
\bottomrule
\end{tabular}
\label{tab:reasoning-effect}
\vspace{-3mm}
\end{table}
\section{Conclusion}

We introduced Omni-MMSI, a new task that requires understanding multi-party social interactions from raw audio-visual input without access to oracle-provided identity-attributed social cues. This setting reflects realistic deployment scenarios where AI systems must operate on automatically extracted cues. To address the resulting challenge of identity attribution, we proposed Omni-MMSI-R, a reference-guided pipeline that aligns multi-modal cues with individual references and performs chain-of-thought (CoT) social reasoning. Through extensive experiments on two social interaction tasks and two social datasets, Omni-MMSI-R demonstrates clear advantages over previous pipelines and advanced Omni-LLMs, achieving state-of-the-art performance. We hope this work establishes a step toward socially intelligent AI that can perceive, reason about, and interact with humans in natural environments.

\section*{Acknowledgements}

We thank Teng Wang for early-stage inspiration that shaped this line of work.
We also thank our colleagues and peers for their valuable feedback and suggestions on this paper.

{
    \small
    \bibliographystyle{ieeenat_fullname}
    \bibliography{main}
}

% \clearpage
% \setcounter{page}{1}
% \maketitlesupplementary

\clearpage
\setcounter{page}{1}

\begin{strip}
\centering
{\Large \bfseries Omni-MMSI: Toward Identity-attributed Social Interaction Understanding\par}
\vspace{0.5em}
{\large \itshape Supplementary Material\par}
\vspace{1.2em}

{
Xinpeng Li$^{1}$ \quad
Bolin Lai$^{2}$ \quad
Hardy Chen$^{3}$ \quad
Shijian Deng$^{1}$ \\
Cihang Xie$^{3}$ \quad
Yuyin Zhou$^{3}$ \quad
James M. Rehg$^{4}$ \quad
Yapeng Tian$^{1}$\par
}

{
$^{1}$University of Texas at Dallas \quad
$^{2}$Georgia Institute of Technology \\
$^{3}$University of California, Santa Cruz \quad
$^{4}$University of Illinois Urbana-Champaign\par
}

{\small
\texttt{\{xinpeng.li, shijian.deng, yapeng.tian\}@utdallas.edu} \\
\small
\texttt{bolin.lai@gatech.edu} \quad 
\small
\texttt{\{hchen403, cixie, yzhou284\}@ucsc.edu} \quad
\small
\texttt{jrehg@illinois.edu} 
}

\end{strip}

\setcounter{section}{0}
\renewcommand{\thesection}{\Alph{section}}
\section{Implementations}

\subsection{Human Study}
\begin{figure}[th]
    \centering
    \includegraphics[width=1.0\linewidth]{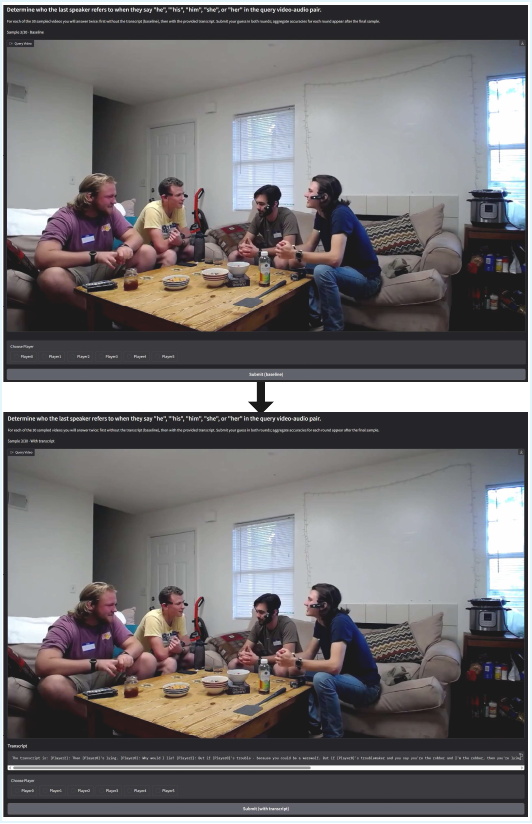}
    \caption{Illustration of human study.}
    \label{fig:supp_human}
    % \vspace{-3mm}
\end{figure}

We randomly selected 30 samples for each participant and first presented the raw video input (Omni-MMSI setting), followed by the version with provided social cues used in the previous setting, as shown in~\cref{fig:supp_human}. This ordering prevents participants from being biased by the provided cues.

\subsection{Human Filtering of CoT Reasoning}

\begin{figure}[h]
    \centering
    \includegraphics[width=1.0\linewidth]{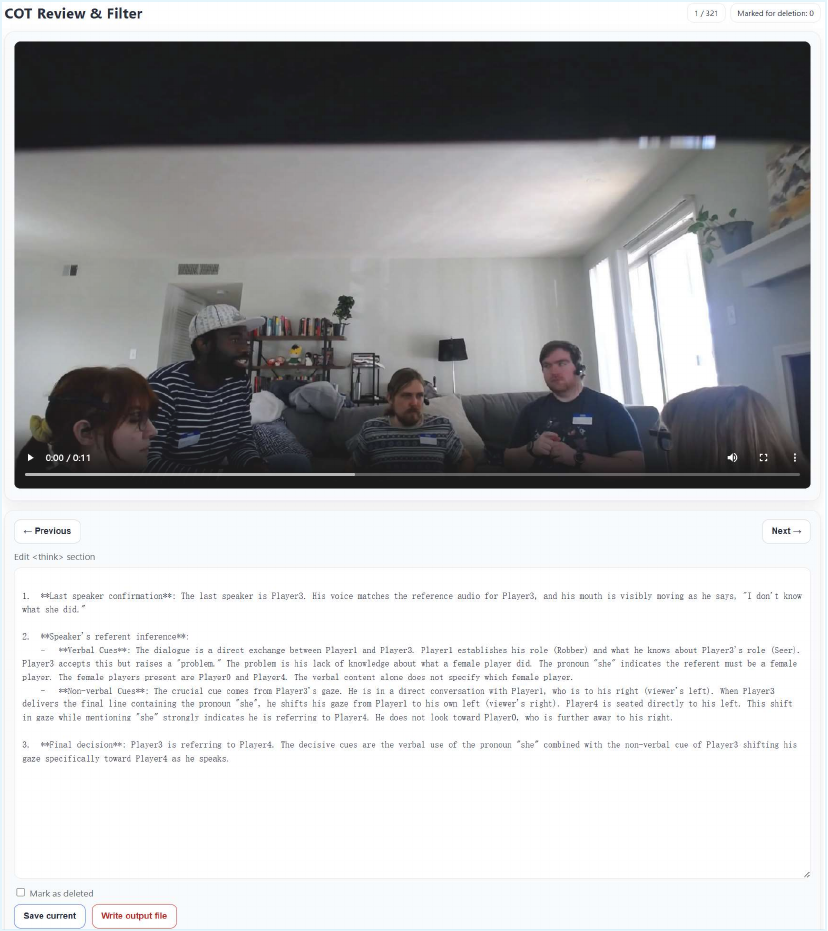}
    \caption{Illustration of human filtering.}
    \label{fig:supp_filter}
    % \vspace{-6mm}
\end{figure}

As shown in \cref{fig:supp_filter}, we examine all reasoning traces that pass the automatic answer-matching step. If a trace contains pervasive errors that fundamentally contradict the audio-visual evidence, we discard it entirely. When only a small number of inaccuracies appear, we manually correct them rather than removing the whole trace. Typical corrections include: a) removing incorrect non-verbal cues, for example, deleting statements such as Player3 looking at Player1 when such gaze does not occur; b) supplementing missing salient evidence, such as adding pointing gestures from the speaker when they serve as a clearer cue than gaze; and c) adding additional non-verbal cues from other participants, for instance, when multiple players are pointing toward the referent but the generated reasoning mentions only the speaker. This process ensures that the final reasoning traces are factually accurate, complete, and faithful.

\subsection{System Prompts}
To generate CoT annotations from the reference-based input, we use the system prompt shown in \Cref{fig:cot_generation_prompt}. This prompt is carefully designed to explicitly instruct the model to identify verbal and non-verbal cues, perform last speaker confirmation and infer the correct referent in a structured step-by-step manner. Its detailed formulation helps the model focus on extracting evidence grounded in the audio-visual input and prevents it from hallucinating unsupported cues. Derived from the CoT-generation prompt, we adopt the system prompt in \Cref{fig:model_training_prompt} for model fine-tuning, the prompt in \Cref{fig:llm_evaluation_prompt_wor} for evaluating Omni-LLMs without references, and the prompt in \Cref{fig:llm_evaluation_prompt_wr} for evaluating Omni-LLMs with references. Overall, these system prompts are not generic instructions; they are deliberately designed and empirically refined to guide the model toward faithful evidence-based reasoning and maximize the effectiveness of reference-based social interaction understanding.

\subsection{System Latency and Parameters}

We report the latency and parameter size of each component in the Omni-MMSI-R pipeline for completeness. All measurements are obtained on an NVIDIA RTX A6000 GPU. For identity-attributed non-verbal cue extraction, YOLO and OSNet together require 0.16s per clip, with 43.69M and 2.17M parameters, respectively. For identity-attributed verbal cue extraction, Whisper and SpeechBrain jointly operate at a 0.21 real-time factor and contain 1541.57M and 22.15M parameters. For the reasoning module, Qwen2.5-Omni (8.93B parameters) produces a direct answer for Omni-MMSI in 1.05s, while enabling chain-of-thought reasoning increases the latency to 12.69s. These numbers characterize the computational profile of the current implementation and serve as a reference for future optimization.

\begin{table*}[t]
\centering
\caption{Latency and parameter size of the components in the Omni-MMSI-R pipeline, measured on an NVIDIA RTX A6000 GPU.}
\label{tab:latency}
\begin{tabular}{lcccc}
\toprule
 & YOLO + OSNet & Whisper + SpeechBrain & Qwen2.5 Omni (Answer) & Qwen2.5 Omni (CoT) \\
\midrule
Latency & 0.16s & 0.21 real-time factor & 1.05s & 12.69s \\
Parameters & 43.69M + 2.17M & 1541.57M + 22.15M & 8.93B & 8.93B \\
\bottomrule
\end{tabular}
% \vspace{-6mm}
\end{table*}

\subsection{Identity Attribution Accuracy Computation}

To compute verbal identity attribution, we first perform sentence-level matching between the predicted utterances $\hat{u}_i$ and the ground-truth utterances $u_i$ using a semantic similarity score. A predicted utterance is considered matched when its similarity exceeds a threshold $\tau_{\text{sem}}{=}0.9$, forming the matched index set $\mathcal{M}_{\text{verb}} = \{\, i \mid \mathrm{sim}(\hat{u}_i, u_i) > \tau_{\text{sem}} \,\}$, where $\mathrm{sim}(\cdot)$ denotes the cosine similarity between sentence embeddings. 
The accuracy is then computed:
\begin{equation}
\mathrm{Acc}_{\text{verb}} =
\frac{1}{|\mathcal{M}_{\text{verb}}|}
\sum_{i \in \mathcal{M}_{\text{verb}}}
\mathds{1}\!\left[\hat{s}_i = s_i\right],
\end{equation}
where $\hat{s}_i$ and $s_i$ represent the predicted and ground-truth speaker identities in the matched pairs, respectively.

For non-verbal identity attribution, we first perform IoU-based matching between the predicted person boxes $\hat{b}_i$ and the ground-truth boxes $b_i$ on the last frame. 
A predicted box is considered matched when its intersection-over-union (IoU) exceeds 
a threshold $\tau_{\text{IoU}}{=}0.9$, forming the matched index set 
$\mathcal{M}_{\text{non-verb}} = \{\, i \mid \mathrm{IoU}(\hat{b}_i, b_i) > \tau_{\text{IoU}} \,\}$. 
The non-verbal attribution accuracy is then computed:
\begin{equation}
\mathrm{Acc}_{\text{non-verb}} =
\frac{1}{|\mathcal{M}_{\text{non-verb}}|}
\sum_{i \in \mathcal{M}_{\text{non-verb}}}
\mathds{1}\!\left[\hat{y}_i = y_i\right],
\end{equation}
where $\hat{y}_i$ and $y_i$ denote the predicted and ground-truth visual identities of the participants, respectively.

\subsection{Task Selection}
We omit Mentioned Player Prediction (MPP) used in prior MMSI~\cite{li2025towards, lee2024modeling}. In the original MMSI formulation, MPP aims to predict the identity referred to by an explicitly mentioned name in a dialogue. The task is constructed by masking a player name (e.g., replacing it with a [MASK] token) and requiring the model to recover the mentioned identity. However, this task is less realistic in practice: AI assistants can typically retrieve explicit names directly from ASR, requiring little social reasoning. Instead, other tasks, STI and PCR, require deeper multimodal cue grounding and social interaction inference. For this reason, we omit MPP in Omni-MMSI and focus on STI and PCR. Since the models are trained and evaluated independently for each task, this omission does not affect comparability with prior works.

\subsection{Task Novelty}
Omni-MMSI is fundamentally different from prior MMSI formulations~\cite{li2025towards, lee2024modeling}. (1) The task assumptions differ. Prior MMSI assumes that identity-attributed social cues are perfectly available, typically via manual annotation or oracle preprocessing.
In contrast, Omni-MMSI requires models to automatically extract identity-attributed cues directly from raw inputs.
(2) The input modality differs.
Previous formulations primarily take visual and textual social cues as input, whereas Omni-MMSI operates on raw multimodal inputs, including visual, text, and audio signals from videos.
Notably, audio is essential for modeling social dynamics such as speaker turns, interruptions, and overlapping speech, which are not supported in prior problems.

\subsection{Reference Reliance}
 When reference information is not pre-stored, the reference bank can be updated automatically. For example, when a person is encountered, the system extracts visual or vocal identity cues, matches them against existing references, and registers a new identity if similarity falls below a threshold. This can be achieved, for example, through a brief greeting-based enrollment step in social scenarios. When references are difficult to obtain (\eg, missing visual), the system degrades to a non-reference mode using raw inputs.

\section{More Results}

\subsection{Robustness of Reference Pairs}
This experiment aims to evaluate the robustness of our reference-based pipeline under audio and visual degradation conditions (on Ego4D), focusing on how noise and occlusion affect verbal and non-verbal attribution accuracy and downstream social interaction understanding tasks.

For audio degradation, we inject additive white Gaussian noise into the reference audio
at signal-to-noise ratio (SNR) levels of \{Clean, 20, 10, 5\}\,dB, where the Clean setting corresponds to no noise injection.
For visual degradation, random occlusion masks are applied to the reference images with occlusion ratios of $\{0.0, 0.1, 0.3, 0.4\}$. The degraded references are used during both the attribution and reasoning stages to assess their overall influence.  

\begin{figure}[th]
    \centering
    \includegraphics[width=0.8\linewidth]{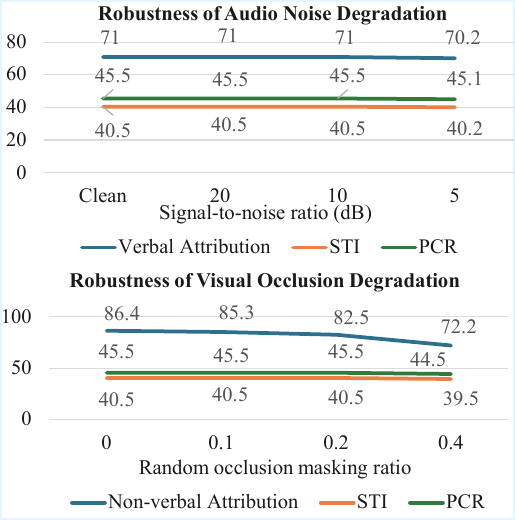}
    \caption{Robustness of the reference-based pipeline under audio and visual degradation. 
(a) Audio noise degradation evaluates the impact of Gaussian noise on verbal attribution and MMSI tasks. 
(b) Visual occlusion degradation tests the effect of partial masking on non-verbal attribution and MMSI tasks.
The results indicate that our pipeline is highly resilient under audio-visual degradation.}
    \label{fig:exp_robust}
    % \vspace{-4mm}
\end{figure}

As shown in \Cref{fig:exp_robust}, decreasing the SNR from 20 to 5 only slightly decreases verbal attribution accuracy from 71.0\% to 70.2\%, with negligible changes in STI and PCR performance (less than 0.5\%). This indicates that the audio branch of our reference-based framework is highly robust to moderate background noise. In contrast, visual degradation results in a moderate performance drop: as the occlusion ratio increases from 0.0 to 0.4 (severe occlusion), non-verbal attribution accuracy decreases from 86.4\% to 72.2\%. Nevertheless, the model maintains stable performance on downstream STI and PCR tasks, showing only marginal variations (around 1\%), demonstrating that high-level social interaction understanding remains robust even under severe visual occlusion. The results indicate that our pipeline remains stable under audio-vision degradation. 

\subsection{Cross-Architecture Generalization}
To further assess robustness of the proposal across scales, we add experiments on Qwen2.5~Omni~3B~\cite{xu2025qwen2} across two datasets and two tasks. From~\cref{tab:3b-comparison}, we observe consistent performance improvements after SFT, both with and without reference inputs. More importantly, incorporating expert tools and CoT reasoning further improves performance. 

\begin{table}[t]
\centering
\caption{Comparison of different settings on Ego4D and YouTube (\%). 
The Omni-LLM backbone is Qwen2.5~Omni~3B.
\textit{ZS}, \textit{SFT}, \textit{ref}, \textit{tool} and \textit{CoT} denote zero-shot inference, supervised fine-tuning, the use of raw reference pairs, tools for extracting identity-attributed social cues and chain-of-thought reasoning supervision. 
}
\begin{tabular}{lccc}
\toprule
Setting (Ego4D) & STI & PCR & Avg. Acc. \\
\midrule
ZS (w/o ref) & 21.14 & 13.61 & 17.38 \\
SFT (w/o ref) & 26.29 & 32.14 & 29.22 \\
ZS (w/ ref) & 21.23 & 10.71 & 15.97 \\
SFT (w/ ref) & 28.57 & 32.85 & 30.71 \\
\rowcolor{gray!15}
SFT (+tool+CoT) & \textbf{33.14} & \textbf{36.56} & \textbf{34.85} \\
\midrule
Setting (YouTube) & STI & PCR & Avg. Acc. \\
\midrule
ZS (w/o ref) & 17.30 & 23.84 & 20.57 \\
SFT (w/o ref) & 23.39 & 25.53 & 24.46 \\
ZS (w/ ref) & 13.61 & 14.97 & 14.29 \\
SFT (w/ ref) & 24.52 & 28.53 & 26.52 \\
\rowcolor{gray!15}
SFT (+tool+CoT) & \textbf{34.88} & \textbf{30.21} & \textbf{32.54} \\
\bottomrule
\end{tabular}
\label{tab:3b-comparison}
% \vspace{-6mm}
\end{table}

\subsection{Additional Comparison on Referential Pipeline}

We tested the zero-shot performance of Omni-LLMs on Ego4D using tool-extracted social cues.
The results show that they remain substantially worse than our SFT model.
This confirms that the observed improvements are not solely due to access to extracted cues, which might contain speech errors.
The performance gain can also arise from effective task formulation and CoT reasoning supervision.

\begin{table}[t]
\centering
\caption{Comparison on Ego4D using tool-extracted social cues. Our model significantly outperforms zero-shot Omni-LLMs, indicating that the gains are not solely from access to extracted cues but also from effective task formulation and CoT supervision.}
\begin{tabular}{lccc}
\toprule
Model & STI & PCR & Avg. Acc. \\
\midrule
OmniVinci & 31.43 & 12.50 & 21.96 \\
Qwen3 Omni 30B & 30.28 & 21.43 & 25.86 \\
Gemini 2.5 Pro & 36.00 & 33.92 & 34.96 \\
\rowcolor{gray!15}
Omni-MMSI-R & \textbf{40.57} & \textbf{45.54} & \textbf{43.06} \\
\bottomrule
\end{tabular}
\vspace{-2mm}
\end{table}

\subsection{Example of CoT Reasoning Trace}
\begin{figure}[th]
    \centering
    \includegraphics[width=1\linewidth]{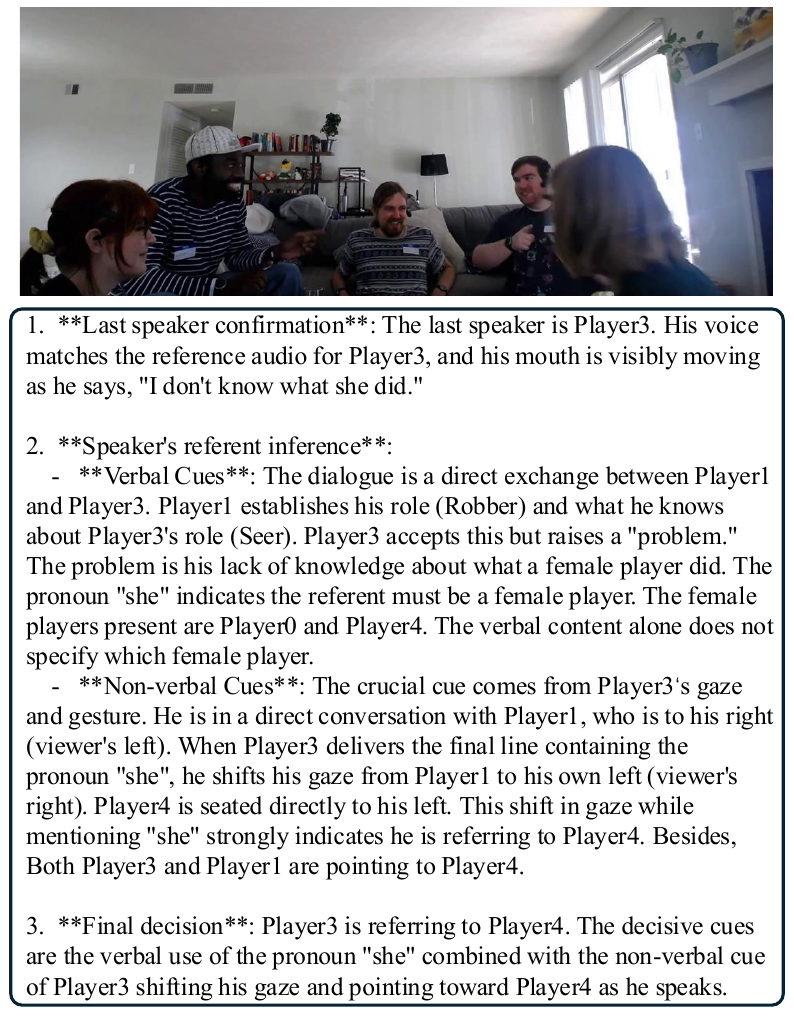}
    \caption{Example of CoT. The CoT performs two key steps-last speaker confirmation and referent inference-to reach the final decision based on reference-based input, leading to reliable prediction.}
    \label{fig:supp_example}
    \vspace{-3mm}
\end{figure}

We present an example of a curated CoT for pronoun coreference recognition. As illustrated in \cref{fig:supp_example}, the CoT performs two key steps-last speaker confirmation and referent inference-to reach the final decision, leading to reliable prediction. The model leverages the identity-attributed transcript and reference audio from the reference-guided input in the last speaker confirmation step. In addition, non-verbal cues such as gaze and gesture are incorporated to complement the verbal evidence in speaker's referent inference. Through supervision from such CoT annotations, the model learns not only structured step-by-step reasoning but also more effective integration of all available social cues.

% \subsection{Failure Cases of Omni-MMSI-R}

\section{Future Works}
Omni-MMSI and Omni-MMSI-R demonstrate promising progress toward identity-attributed social interaction understanding, which better supports future exploration of richer social scenes and social tasks. A current limitation is that the datasets used in this work represent controlled scenarios where all participants remain visible under a fixed game setting. Although such setups make manually identity attribution easy, they capture only a narrow portion of real-world social dynamics. In natural environments, people enter or exit the scene and camera viewpoints often change abruptly. Multi-person interactions in movies, television content, and outdoor gatherings also involve frequent camera cuts and heterogeneous visual contexts. Under our reference-based design, these multi-shot and camera-switching scenes, which were previously difficult to study because prior methods could not maintain consistent attribution across shots, become substantially more feasible to curate. Stable reference identities allow reliable cross-shot identity grounding, enabling richer and more realistic social scenarios to be included in future datasets. Extending Omni-MMSI to larger and more diverse environments is an important direction for improving real-world applicability.

\section{Societal Impacts and Concerns}
While Omni-MMSI aims to advance socially-intelligent AI assistants by enabling real-world perception and reasoning over individual-level verbal and non-verbal cues, these same capabilities introduce potential societal risks. In particular, the ability to align speech, gaze, and gestures with specific individuals, central to our reference-based cue attribution framework, could be misused for intrusive monitoring of social interactions, workplace surveillance, or targeted behavioral manipulation if deployed without consent or appropriate safeguards. Moreover, because Omni-MMSI operates on imperfectly extracted audio-visual cues, systematic errors in speech recognition, tracking, or gaze estimation may disproportionately affect certain demographic groups, potentially amplifying existing biases in downstream decisions. These concerns highlight that the contributions in this work are intended strictly for research on accurate multi-modal social understanding rather than for surveillance applications. Responsible deployment of such systems requires strong privacy protections, transparent usage policies, and governance mechanisms that prevent misuse, especially in real-world settings where individual-level attribution carries heightened ethical implications.

\begin{figure*}[htbp]
\centering
\footnotesize
\begin{AIbox}{System Prompt for CoT Generation}
% {\bf Prompt:} \\
{\footnotesize
\# Inputs 

\#\# Reference image-audio pairs \\
You are provided with a reference image-audio pair for each player (referred to as PlayerN). 
These references are provided to help the model extract attributed verbal and non-verbal social cues by aligning and comparing them with the reference image-audio pairs, enabling the alignment of social cues. \\

\#\# Bounding box coordinates \\
Each player is associated with a bounding box in the format: 
\texttt{\string[PlayerN\string]}: [$x_{min}, y_{min}, x_{max}, y_{max}$]. 
These coordinates are preliminary and may contain errors or drift. Use them to locate players in the query video to help reason about who is speaking and who the speaker is referring to. \\

\#\# Transcription \\
You are provided with a transcription with several utterance segments for the query video-audio pair in the format: \texttt{\string[PlayerN\string]}: [utterance content]. 
These segments are preliminary and may contain incorrect speaker tags or wording. Use them to help reason about who is speaking and who the speaker is referring to. You can transcribe the target audio and use commonsense to correct preliminary transcript. \\

\#\# Query video-audio pair \\
This is the video-audio pair in which you must extract attributed verbal and non-verbal social cues and determine who the last speaker is referring to. \\
 \\
\# Task \\
Determine **who the last speaker refers to when they say ``he'', ``his'', ``him'', ``she'', or ``her''** in the query video-audio pair. \\
Treat the task as two mandatory stages: \\
1. Extract attributed verbal and non-verbal social cues, including speaker-attributed transcript and speaker-attributed visual behaviors like gaze and gesture. \\
2. Analyze these social cues to infer the last speaker and their referent, especially speech content, dialog turn-taking, and visual engagement. \\
 \\
\# Decision rules \\
1. Denote speakers as Player0, Player1, etc., based on their position from left to right in the video. \\
2. Referent must be a PlayerN, including off-screen or occluded Players that are included in conversation. \\
3. Every conclusion must cite both **verbal** signals (e.g., speaker content matching with previous dialog content, previous speaker) and **non-verbal** cues (e.g., speaker and who are making mutual eye contact, speaker is pointing at whom). Generic statements like ``they are near each other'' without interaction detail are insufficient. \\
 \\
\# Required social cues structure \\
Inside \textless cue\textgreater ... \textless /cue\textgreater\ explicitly cover: \\
- Speaker-attributed transcript. Each utterance represents one utterance segment in chronological order. \\
- Speaker-attributed visual behaviors. You must include the bounding box coordinates [x\_min, y\_min, x\_max, y\_max] for each player when describing their visual behaviors. \\
 \\
\# Required reasoning structure \\
Inside \textless think\textgreater ... \textless /think\textgreater\ explicitly cover: \\
1. Last speaker confirmation with provided audio, vision, and speech evidence. \\
2. Speaker's referent inference from the verbal and non-verbal interaction cues. \\
3. Final decision that names the referent and states the decisive cues. \\
 \\
\# Output format \\
1. A social cues trace wrapped in \textless cue\textgreater ... \textless /cue\textgreater\ that follows the structure above. \\
2. A reasoning trace wrapped in \textless think\textgreater ... \textless /think\textgreater\ that follows the structure above. \\
3. The final answer wrapped in \textless answer\textgreater PlayerN \textless /answer\textgreater. \\
 \\
\# Examples \\
\textless cue\textgreater 
The verbal cues of all players are: 
\texttt{\string[Player1\string]}: And then this- 
\texttt{\string[Player2\string]}: That one was you three. 
\texttt{\string[Player1\string]}: Yeah. 
\texttt{\string[Player3\string]}: So it's between us two? 
\texttt{\string[Player0\string]}: Yeah. 
\texttt{\string[Player3\string]}: You said, ``If you keep going on the rejection strategy you'll lose,'' meaning he's not on the good team, because I am.'' 
The non-verbal cues of all players are:  
Player0 ([0.001, 0.719, 0.165, 0.992]) is looking at Player4's hands.  
Player1 ([0.125, 0.597, 0.333, 0.992]) glances at the cards in their hand, then looks up.
Player2 ([0.379, 0.728, 0.519, 0.992]) has a look at his cards on the table and then looks at Player4. 
Player3 ([0.546, 0.664, 0.703, 0.989]) is looking at his watch. 
Player4 ([0.679, 0.728, 0.872, 0.989]) is only visible from the left rear side, but it can be inferred that she looks at the cards on the table and then faces Player2 directly. 
\textless /cue\textgreater 
\textless think\textgreater 
1. Last speaker: Player4 delivers the final utterance ``Okay. Do you need the script?'' and their voice matches the Player4 reference while the Player4 bounding box shows their mouth moving.  
2. Speaker referents: Player2 (responded just before), Player3 (standing nearby but disengaged).  
3. Decision: Player4 addresses Player2 based on directed gesture and mutual gaze. 
\textless /think\textgreater 
\textless answer\textgreater Player2 \textless /answer\textgreater 

}
\end{AIbox} 
\caption{System prompt for CoT generation. }
\label{fig:cot_generation_prompt}
\end{figure*}

\begin{figure*}[htbp]
\centering
\footnotesize
\begin{AIbox}{System Prompt for Model Training}
% {\bf Prompt:} \\
{\footnotesize
\# Inputs 

\#\# Reference image-audio pairs \\
You are provided with a reference image-audio pair for each player (referred to as PlayerN). 
These references are provided to help the model extract attributed verbal and non-verbal social cues by aligning and comparing them with the reference image-audio pairs, enabling the alignment of social cues. \\

\#\# Bounding box coordinates \\
Each player is associated with a bounding box in the format: 
\texttt{\string[PlayerN\string]}: [$x_{min}, y_{min}, x_{max}, y_{max}$]. 
These coordinates are preliminary and may contain errors or drift. Use them to locate players in the query video to help reason about who is speaking and who the speaker is referring to. \\

\#\# Transcription \\
You are provided with a transcription with several utterance segments for the query video-audio pair in the format: \texttt{\string[PlayerN\string]}: [utterance content]. 
These segments are preliminary and may contain incorrect speaker tags or wording. Use them to help reason about who is speaking and who the speaker is referring to. You can transcribe the target audio and use commonsense to correct preliminary transcript. \\

\#\# Query video-audio pair \\
This is the video-audio pair in which you must extract attributed verbal and non-verbal social cues and determine who the last speaker is referring to. \\
 \\
\# Task \\
Determine **who the last speaker refers to when they say ``he'', ``his'', ``him'', ``she'', or ``her''** in the query video-audio pair. \\
Treat the task as two mandatory stages: \\
1. Extract attributed verbal and non-verbal social cues, including speaker-attributed transcript and speaker-attributed visual behaviors like gaze and gesture. \\
2. Analyze these social cues to infer the last speaker and their referent, especially speech content, dialog turn-taking, and visual engagement. \\
 \\
\# Decision rules \\
1. Denote speakers as Player0, Player1, etc., based on their position from left to right in the video. \\
2. Referent must be a PlayerN, including off-screen or occluded Players that are included in conversation. \\
3. Every conclusion must cite both **verbal** signals (e.g., speaker content matching with previous dialog content, previous speaker) and **non-verbal** cues (e.g., speaker and who are making mutual eye contact, speaker is pointing at whom). Generic statements like ``they are near each other'' without interaction detail are insufficient. \\
 \\
\# Required reasoning structure \\
Inside \textless think\textgreater ... \textless /think\textgreater\ explicitly cover: \\
1. Last speaker confirmation with provided audio, vision, and speech evidence. \\
2. Speaker's referent inference from the verbal and non-verbal interaction cues. \\
3. Final decision that names the referent and states the decisive cues. \\
 \\
\# Output format \\
1. A reasoning trace wrapped in \textless think\textgreater ... \textless /think\textgreater\ that follows the structure above. \\
2. The final answer wrapped in \textless answer\textgreater PlayerN \textless /answer\textgreater. \\
 \\
\# Examples \\
\textless think\textgreater 
1. Last speaker: Player4 delivers the final utterance ``Okay. Do you need the script?'' and their voice matches the Player4 reference while the Player4 bounding box shows their mouth moving.  
2. Speaker referents: Player2 (responded just before), Player3 (standing nearby but disengaged).  
3. Decision: Player4 addresses Player2 based on directed gesture and mutual gaze. 
\textless /think\textgreater 
\textless answer\textgreater Player2 \textless /answer\textgreater 

}
\end{AIbox} 
\caption{System prompt for model training. }
\label{fig:model_training_prompt}
\end{figure*}

\begin{figure*}[htbp]
\centering
\footnotesize
\begin{AIbox}{System Prompt for Omni-LLMs Evaluation w/o Reference}
% {\bf Prompt:} \\
{\footnotesize
\# Inputs 

\#\# Query video-audio pair \\
This is the video-audio pair in which you must extract attributed verbal and non-verbal social cues and determine who the last speaker is referring to. \\
 \\
\# Task \\
Determine **who the last speaker refers to when they say ``he'', ``his'', ``him'', ``she'', or ``her''** in the query video-audio pair. \\
Treat the task as two mandatory stages: \\
1. Extract attributed verbal and non-verbal social cues, including speaker-attributed transcript and speaker-attributed visual behaviors like gaze and gesture. \\
2. Analyze these social cues to infer the last speaker and their referent, especially speech content, dialog turn-taking, and visual engagement. \\
 \\
\# Decision rules \\
1. Denote speakers as Player0, Player1, etc., based on their position from left to right in the video. \\
2. Referent must be a PlayerN, including off-screen or occluded Players that are included in conversation. \\
3. Every conclusion must cite both **verbal** signals (e.g., speaker content matching with previous dialog content, previous speaker) and **non-verbal** cues (e.g., speaker and who are making mutual eye contact, speaker is pointing at whom). Generic statements like ``they are near each other'' without interaction detail are insufficient. \\
 \\
\# Required social cues structure \\
Inside \textless cue\textgreater ... \textless /cue\textgreater\ explicitly cover: \\
- Speaker-attributed transcript. Each utterance represents one utterance segment in chronological order. \\
- Speaker-attributed visual behaviors. You must include the bounding box coordinates [x\_min, y\_min, x\_max, y\_max] for each player when describing their visual behaviors. \\
 \\
\# Required reasoning structure \\
Inside \textless think\textgreater ... \textless /think\textgreater\ explicitly cover: \\
1. Last speaker confirmation with provided audio, vision, and speech evidence. \\
2. Speaker's referent inference from the verbal and non-verbal interaction cues. \\
3. Final decision that names the referent and states the decisive cues. \\
 \\
\# Output format \\
1. A social cues trace wrapped in \textless cue\textgreater ... \textless /cue\textgreater\ that follows the structure above. \\
2. A reasoning trace wrapped in \textless think\textgreater ... \textless /think\textgreater\ that follows the structure above. \\
3. The final answer wrapped in \textless answer\textgreater PlayerN \textless /answer\textgreater. \\
 \\
\# Examples \\
\textless cue\textgreater 
The verbal cues of all players are: 
\texttt{\string[Player1\string]}: And then this- 
\texttt{\string[Player2\string]}: That one was you three. 
\texttt{\string[Player1\string]}: Yeah. 
\texttt{\string[Player3\string]}: So it's between us two? 
\texttt{\string[Player0\string]}: Yeah. 
\texttt{\string[Player3\string]}: You said, ``If you keep going on the rejection strategy you'll lose,'' meaning he's not on the good team, because I am.'' 
The non-verbal cues of all players are:  
Player0 ([0.001, 0.719, 0.165, 0.992]) is looking at Player4's hands.  
Player1 ([0.125, 0.597, 0.333, 0.992]) glances at the cards in their hand, then looks up.
Player2 ([0.379, 0.728, 0.519, 0.992]) has a look at his cards on the table and then looks at Player4. 
Player3 ([0.546, 0.664, 0.703, 0.989]) is looking at his watch. 
Player4 ([0.679, 0.728, 0.872, 0.989]) is only visible from the left rear side, but it can be inferred that she looks at the cards on the table and then faces Player2 directly. 
\textless /cue\textgreater 
\textless think\textgreater 
1. Last speaker: Player4 delivers the final utterance ``Okay. Do you need the script?'' and their voice matches the Player4 reference while the Player4 bounding box shows their mouth moving.  
2. Speaker referents: Player2 (responded just before), Player3 (standing nearby but disengaged).  
3. Decision: Player4 addresses Player2 based on directed gesture and mutual gaze. 
\textless /think\textgreater 
\textless answer\textgreater Player2 \textless /answer\textgreater 

}
\end{AIbox} 
\caption{System prompt for Omni-LLMs evaluation without reference.}
\label{fig:llm_evaluation_prompt_wor}
\end{figure*}

\begin{figure*}[htbp]
\centering
\footnotesize
\begin{AIbox}{System Prompt for Omni-LLMs Evaluation w/ Reference}
% {\bf Prompt:} \\
{\footnotesize
\# Inputs 

\#\# Reference image-audio pairs \\
You are provided with a reference image-audio pair for each player (referred to as PlayerN). 
These references are provided to help the model extract attributed verbal and non-verbal social cues by aligning and comparing them with the reference image-audio pairs, enabling the alignment of social cues. \\

\#\# Query video-audio pair \\
This is the video-audio pair in which you must extract attributed verbal and non-verbal social cues and determine who the last speaker is referring to. \\
 \\
\# Task \\
Determine **who the last speaker refers to when they say ``he'', ``his'', ``him'', ``she'', or ``her''** in the query video-audio pair. \\
Treat the task as two mandatory stages: \\
1. Extract attributed verbal and non-verbal social cues, including speaker-attributed transcript and speaker-attributed visual behaviors like gaze and gesture. \\
2. Analyze these social cues to infer the last speaker and their referent, especially speech content, dialog turn-taking, and visual engagement. \\
 \\
\# Decision rules \\
1. Denote speakers as Player0, Player1, etc., based on their position from left to right in the video. \\
2. Referent must be a PlayerN, including off-screen or occluded Players that are included in conversation. \\
3. Every conclusion must cite both **verbal** signals (e.g., speaker content matching with previous dialog content, previous speaker) and **non-verbal** cues (e.g., speaker and who are making mutual eye contact, speaker is pointing at whom). Generic statements like ``they are near each other'' without interaction detail are insufficient. \\
 \\
\# Required social cues structure \\
Inside \textless cue\textgreater ... \textless /cue\textgreater\ explicitly cover: \\
- Speaker-attributed transcript. Each utterance represents one utterance segment in chronological order. \\
- Speaker-attributed visual behaviors. You must include the bounding box coordinates [x\_min, y\_min, x\_max, y\_max] for each player when describing their visual behaviors. \\
 \\
\# Required reasoning structure \\
Inside \textless think\textgreater ... \textless /think\textgreater\ explicitly cover: \\
1. Last speaker confirmation with provided audio, vision, and speech evidence. \\
2. Speaker's referent inference from the verbal and non-verbal interaction cues. \\
3. Final decision that names the referent and states the decisive cues. \\
 \\
\# Output format \\
1. A social cues trace wrapped in \textless cue\textgreater ... \textless /cue\textgreater\ that follows the structure above. \\
2. A reasoning trace wrapped in \textless think\textgreater ... \textless /think\textgreater\ that follows the structure above. \\
3. The final answer wrapped in \textless answer\textgreater PlayerN \textless /answer\textgreater. \\
 \\
\# Examples \\
\textless cue\textgreater 
The verbal cues of all players are: 
\texttt{\string[Player1\string]}: And then this- 
\texttt{\string[Player2\string]}: That one was you three. 
\texttt{\string[Player1\string]}: Yeah. 
\texttt{\string[Player3\string]}: So it's between us two? 
\texttt{\string[Player0\string]}: Yeah. 
\texttt{\string[Player3\string]}: You said, ``If you keep going on the rejection strategy you'll lose,'' meaning he's not on the good team, because I am.'' 
The non-verbal cues of all players are:  
Player0 ([0.001, 0.719, 0.165, 0.992]) is looking at Player4's hands.  
Player1 ([0.125, 0.597, 0.333, 0.992]) glances at the cards in their hand, then looks up.
Player2 ([0.379, 0.728, 0.519, 0.992]) has a look at his cards on the table and then looks at Player4. 
Player3 ([0.546, 0.664, 0.703, 0.989]) is looking at his watch. 
Player4 ([0.679, 0.728, 0.872, 0.989]) is only visible from the left rear side, but it can be inferred that she looks at the cards on the table and then faces Player2 directly. 
\textless /cue\textgreater 
\textless think\textgreater 
1. Last speaker: Player4 delivers the final utterance ``Okay. Do you need the script?'' and their voice matches the Player4 reference while the Player4 bounding box shows their mouth moving.  
2. Speaker referents: Player2 (responded just before), Player3 (standing nearby but disengaged).  
3. Decision: Player4 addresses Player2 based on directed gesture and mutual gaze. 
\textless /think\textgreater 
\textless answer\textgreater Player2 \textless /answer\textgreater 

}
\end{AIbox} 
\caption{System prompt for Omni-LLMs evaluation with reference}
\label{fig:llm_evaluation_prompt_wr}
\end{figure*}

\end{document}